# Cyberbullying Detection in Social Networks Using Deep Learning Based Models; A Reproducibility Study

Maral Dadvar and Kai Eckert

Web-based Information Systems and Services, Stuttgart Media University
Nebenstrasse 8, 70569 Stuttgart, Germany
`{dadvar , eckert}@hdm-stuttgart.de`

**Abstract.** Cyberbullying is a disturbing online misbehaviour with troubling consequences. It appears in different forms, and in most of the social networks, it is in textual format. Automatic detection of such incidents requires intelligent systems. Most of the existing studies have approached this problem with conventional machine learning models and the majority of the developed models in these studies are adaptable to a single social network at a time. In recent studies, deep learning based models have found their way in the detection of cyberbullying incidents, claiming that they can overcome the limitations of the conventional models, and improve the detection performance. In this paper, we investigate the findings of a recent literature in this regard. We successfully reproduced the findings of this literature and validated their findings using the same datasets, namely Wikipedia, Twitter, and Formspring, used by the authors. Then we expanded our work by applying the developed methods on a new YouTube dataset (~54k posts by ~4k users) and investigated the performance of the models in new social media platforms. We also transferred and evaluated the performance of the models trained on one platform to another platform. Our findings show that the deep learning based models outperform the machine learning models previously applied to the same YouTube dataset. We believe that the deep learning based models can also benefit from integrating other sources of information and looking into the impact of profile information of the users in social networks.

**Keywords:** Deep Learning, Online Bullying, Neural Networks, Social Networks, Transfer Learning, YouTube.

## 1      Introduction

With the emergence of Web 2.0 there has been a substantial impact on social communication, and relationships and friendships have been redefined all over again. Adolescents spend a considerable amount of time online and on different social platforms, besides all the benefits that it might bring them, their online presence also make them vulnerable to threats and social misbehaviours such as cyberbullying.

Studies show that about 18% of the children in Europe have been involved in cyberbullying[1]. In the 2014 EU Kids Online Report [1] it is stated that 20% of children from 11 to 16 years old have been exposed to online bullying. The quantitative research of [2] shows that cyber-victimization rates among teenagers between 20% and 40%. These statistics go on and on[2,3] [3]. All these demonstrate the importance of finding a robust and comprehensive solution to this widespread problem.

Cyberbullying needs to be understood and addressed from different perspectives. Automatic detection and prevention of these incidents can substantially help to tackle this problem. There are already tools developed which can flag a bullying incident [4] and programs which try to provide support to the victims [5]. Moreover, most of the online platforms which are commonly used by teenagers have safely centres, for example, YouTube Safety Centre[4] and Twitter Safety and Security[5], which provide support to users and monitor the communications. There are also many research conducted on automatic detection and prevention of cyberbullying, which we will address in more details in the next section, but this problem is still far from resolved and there is the need for further improvements towards having a concrete solution. Most of the existing studies [6]–[9] have used conventional Machine Learning (ML) models to detect cyberbullying incidents. Recently Deep Neural Network Based (DNN) models have also been applied for detection of cyberbullying [10], [11].

In [11], authors have used DNN models for detection of cyberbullying and have expanded their models across multiple social media platforms. Based on their reported results, their models outperform traditional ML models, and most importantly authors have stated that they have applied transfer learning which means their developed models for detection of cyberbullying can be adapted and used on other datasets.

In this contribution, we begin by reproducing and validating the [11] proposed models and their results on the three datasets, Formspring [12], Twitter [13] and Wikipedia [14], which have been used by the authors. Cyberbullying takes place in almost all of the online social networks; therefore, developing a detection model which is adaptable and transferable to different social networks is of great value. We expand our work by re-implementing the models on a new dataset. For this purpose, we have used a YouTube dataset which has been extensively used in cyberbullying studies [6], [15], [16]. The ultimate aim was to investigate the interoperability and the performance of the reproduced models on new datasets, to see how adaptable they are to different social media platforms and to what extent models trained on a dataset (i.e., social network) can be transferred to another one. This provides a base to compare the outcome of DNN models with the conventional ML models. In the remainder of this paper, we reported the reproduced experimental setup, datasets and findings (Section 2), we investigated the adaptability of the methods on the YouTube dataset (Section 3), and in Section 4 we discussed our findings, compared our results with previous attempts on the same dataset, and pointed out potential future works.

---

[1] EU COST Action IS0801on Cyberbullying (https://sites.google.com/site/costis0801/)
[2] http://www.ditchthelabel.org/annual-cyber-bullying-survey-cyber-bullying-statistics
[3] https://www.ncpc.org
[4] https://www.youtube.com/yt/about/policies/#staying-safe
[5] https://help.twitter.com/en/safety-and-security

## 2 Reproduced Experimental Setup

In this study, we have first reproduced the experiments conducted in [11] on the datasets used by the authors namely, Formspring [12], Wikipedia [14], and Twitter [13]. We have used the same models and experimental setup for our implementations. In this section, we have briefly introduced the datasets and explained the models and other experiment components. For further details please see the reference literature.

### 2.1 Datasets

In this study three datasets are used; Formspring (a Q&A forum), Wikipedia talk pages (collaborative knowledge repository) and Twitter (microblogging platform). All these datasets are manually labelled and publicly available. One problem that all these datasets share, is that the datasets are skewed, and there is a class imbalance, i.e., the number of posts labelled as bullying is significantly less than the neutral ones. This causes the classification results to be biased towards non-bullying posts. Therefore as it will be explained in more details in the next section, we oversampled the bullying class in the datasets. Furthermore, the size of the posts in terms of number of words differs across datasets. This can affect the number of distinct words encountered in each dataset. Therefore, long posts (measured based on the number of words) are truncated to the size of post ranked at 95 percentile in that dataset.

**Wikipedia.** A talk page in Wikipedia is where the discussions among those who have contributed to the editing of a Wikipedia page are maintained. This dataset includes more than 10,000 discussion comments from English Wikipedia' talk pages. All comments were manually annotated by ten persons. In total 13,590 comments were labelled as a personal attack.

**Formspring.** This is a question-answering platform. The dataset includes 12,000 pairs of questions-answers which were manually annotated by three persons. In total 825 pairs was annotated as cyberbullying by at least two persons.

**Twitter.** From this microblogging platform, 16,000 tweets were collected and manually annotated. The tweets were collected by search of terms which refer to religious, sexual, gender, and ethnic minorities. In total 3117 were labelled as sexist and 1937 were labelled as racist. The remaining tweets were labelled as neither.

## 2.2   Models and Methods

In this section the four DDN models that we experimented for cyberbullying detection are described. We also have experimented with three methods for initializing word embeddings, which will be briefly explained in follow.

**Deep Neural Network Based Models.** In this study four different DNN models were used for detection of cyberbullying: Convolutional Neural Network (CNN), Long Short-Term Memory (LSTM), Bidirectional LSTM (BLSTM) and BLSTM with attention. These models respectively vary in complexity in their neural architecture. CNNs are mostly used for image and text classification [17], [18] as well as sentiment classification [19]. LSTM networks are used for learning long-term dependencies. Their internal memory makes these networks useful for text classification [20], [21]. Bidirectional LSTMs, increase the input information to the network by encoding information in both forward and backward direction [22]. BLSTM with attention, gives more direct dependence between the state of the model at different points in time [23].

All the models have identical layers except for the neural architecture layer which is unique to each model. The embedding layer, which will be explained in more details in following, processes a fixed length sequence of words. There are two dropout layers which are used to avoid overfitting, once before (with dropout rates of 0.25) and one after (with dropout rates of 0.5) the neural architecture layer. Then there is the fully connected layer which is a dense output layer with the number of neurons equal to the number of classes. The last layer is the softmax layer that provides softmax activation. All the models are trained using backpropagation with Adam optimizer and categorical cross-entropy loss function. All the reference codes can be found in the author's GitHub repository[6].

**Initial Word Embedding.** Word embedding is the process of representing each word as real value vector. The embedding layer of our models processes a fixed length sequence of words. In this study three methods are used for initializing word embeddings: random, GloVe [24] and SSWE [25]. Using initial words embeddings during the training can improve the model to learn task specific word embeddings. Task specific word embeddings can differentiate the style of cyberbullying among different online platform as well as topics. The GloVe vectors mostly improve the performance of the models over the random vector initialization. However, in GloVe method only the syntactic context of the word in considered and the sentiment conveyed by the text is ignored. This problem is overcome in the SSWE method by incorporating the text sentiment as one of the parameters for word embedding generation. In this study different dimension size for word embeddings are experimented. Since there was no significant difference in the result of dimensions from 30 to 200, the results of the dimension size 50 are reported.

---

[6]   https://github.com/sweta20/Detecting-Cyberbullying-Across-SMPs

## 2.3    Workflow and Results

We started our experiments by implementing the DDN based models using Keras Python package[7]. The datasets went through the pre-processing steps such as stop-words and punctuations removal. The performance of the models were evaluated using five-fold cross-validation and precision, recall and *F1*-score evaluation metrics were used to report the performance. In order to overcome the problem caused by the imbalance in the datasets, the number of bullying class in the datasets were oversampled and the bullying posts were tripled in the training data. To illustrate the effect of oversampling, the BLSTM with attention model was used with three word embedding methods, once on the original datasets and once on the oversampled datasets (see Table 1a).

**Table 1a.** Performance evaluation of the BLSTM with attention models, with three word embeddings on the original and the oversamples datasets.

| Dataset | Label | Precision | | | Recall | | | *F1*-score | | |
|---|---|---|---|---|---|---|---|---|---|---|
| | | Random | Glove | SSWE | Random | Glove | SSWE | Random | Glove | SSWE |
| **F** | Bully | 0.37 | 0.60 | 0.50 | 0.34 | 0.43 | 0.43 | 0.35 | 0.50 | 0.46 |
| **F⁺** | Bully | 0.91 | 0.88 | 0.86 | 0.98 | 0.95 | 0.92 | 0.94 | 0.91 | 0.89 |
| **T** | Racism | 0.81 | 0.84 | 0.82 | 0.71 | 0.69 | 0.75 | 0.76 | 0.76 | 0.78 |
| **T⁺** | Racism | 0.95 | 0.95 | 0.96 | 0.99 | 0.99 | 0.99 | 0.97 | 0.97 | 0.97 |
| **T** | Sexism | 0.73 | 0.75 | 0.71 | 0.71 | 0.85 | 0.77 | 0.72 | 0.79 | 0.74 |
| **T⁺** | Sexism | 0.95 | 0.94 | 0.94 | 0.99 | 0.99 | 0.99 | 0.97 | 0.96 | 0.97 |
| **W** | Attack | 0.80 | 0.76 | 0.81 | 0.68 | 0.74 | 0.65 | 0.74 | 0.75 | 0.72 |
| **W⁺** | Attack | 0.94 | 0.93 | 0.92 | 0.99 | 0.98 | 0.98 | 0.96 | 0.96 | 0.95 |

F: Formspring, T: Twitter, W: Wikipedia, ⁺: Oversampled

**Table 1b.** Difference in performance of the reference and the reproduced BLSTM with attention models, with three word embeddings on the original and the oversamples datasets. Negative values indicate higher performance of the reproduced models.

| Dataset | Label | Precision | | | Recall | | | *F1*-score | | |
|---|---|---|---|---|---|---|---|---|---|---|
| | | Random | Glove | SSWE | Random | Glove | SSWE | Random | Glove | SSWE |
| **F** | Bully | 0.15 | -0.04 | 0.13 | 0.06 | 0.06 | -0.05 | 0.09 | 0.01 | 0.01 |
| **F⁺** | Bully | -0.07 | -0.03 | 0.04 | 0.00 | 0.02 | -0.01 | -0.04 | -0.01 | 0.02 |
| **T** | Racism | -0.14 | -0.10 | -0.06 | 0.02 | 0.07 | 0.02 | -0.06 | -0.01 | -0.02 |
| **T⁺** | Racism | -0.01 | -0.05 | -0.06 | -0.01 | -0.04 | -0.03 | -0.01 | -0.04 | -0.04 |
| **T** | Sexism | -0.08 | 0.11 | 0.12 | -0.07 | -0.33 | -0.30 | -0.07 | -0.14 | -0.15 |

---

[7]    https://keras.io/

| | | | | | | | | | |
|---|---|---|---|---|---|---|---|---|---|
| **T**[+] | Sexism | -0.07 | 0.01 | -0.06 | -0.02 | -0.08 | -0.07 | -0.04 | -0.05 | -0.07 |
| **W** | Attack | -0.03 | 0.05 | 0.01 | 0.06 | -0.07 | 0.03 | 0.02 | -0.01 | 0.02 |
| **W**[+] | Attack | -0.13 | -0.07 | -0.05 | -0.08 | -0.09 | -0.12 | -0.08 | -0.08 | -0.08 |

F: Formspring, T: Twitter, W: Wikipedia, [+]: Oversampled

The oversampling has significantly (Mann-Whitney U test, P<0.001) improved the performance of the models in all the datasets, especially those with a smaller number of bullying posts in their training data such as Formspring. Initial word embeddings affect the data representation for the models which use them during the training to learn task specific word embeddings. Comparing the performance of the reference and the reproduced models shows that the majority of the reproduced results were within the standard deviation of the reference results (Table 1b). The highest inconsistencies were observed in the recall values of original Twitter dataset with GloVe and SSWE word embeddings.

In Table 2a we have reported the F1-scores of different initial word embeddings on two DDN models; CNN as the simplest model, and BLSTM with attention as the most complex model.

**Table 2a.** Comparison of the *F1*-scores for CNN and BLSTM with attention models with three word embedding methods on the original and the oversamples datasets.

| Dataset | Label | Random | | GloVe | | SSWE | |
|---|---|---|---|---|---|---|---|
| | | CNN | BLSTM attention | CNN | BLSTM attention | CNN | BLSTM attention |
| **F** | Bully | 0.41 | 0.35 | 0.33 | 0.5 | 0.27 | 0.46 |
| **F**[+] | Bully | 0.89 | 0.94 | 0.86 | 0.91 | 0.74 | 0.89 |
| **T** | Racism | 0.77 | 0.76 | 0.68 | 0.76 | 0.66 | 0.78 |
| **T**[+] | Racism | 0.90 | 0.97 | 0.90 | 0.97 | 0.86 | 0.97 |
| **T** | Sexism | 0.58 | 0.72 | 0.46 | 0.79 | 0.48 | 0.74 |
| **T**[+] | Sexism | 0.79 | 0.97 | 0.80 | 0.96 | 0.73 | 0.97 |
| **W** | Attack | 0.71 | 0.74 | 0.72 | 0.75 | 0.67 | 0.72 |
| **W**[+] | Attack | 0.87 | 0.96 | 0.84 | 0.96 | 0.87 | 0.95 |

F: Formspring, T: Twitter, W: Wikipedia, [+]: Oversampled

**Table 2b.** Difference in the *F1*-scores of the reference and the reproduced CNN and BLSTM with attention models with three word embedding methods on the original and the oversamples datasets. Negative values indicate higher performance of the reproduced models.

| Dataset | Label | Random | | GloVe | | SSWE | |
|---|---|---|---|---|---|---|---|
| | | CNN | BLSTM attention | CNN | BLSTM attention | CNN | BLSTM attention |
| **F** | Bully | -0.11 | 0.09 | 0.01 | 0.01 | 0.07 | 0.01 |
| **F⁺** | Bully  | 0.02  | -0.04 | 0.07 | -0.01 | 0.17 | 0.02  |
| **T**  | Racism | -0.09 | -0.06 | 0.05 | -0.01 | 0.04 | -0.02 |
| **T⁺** | Racism | 0.00  | -0.01 | 0.05 | -0.04 | 0.07 | -0.04 |
| **T**  | Sexism | 0.01  | -0.07 | 0.15 | -0.14 | 0.15 | -0.15 |
| **T⁺** | Sexism | 0.14  | -0.04 | 0.13 | -0.05 | 0.19 | -0.07 |
| **W**  | Attack | 0.01  | 0.02  | 0.00 | -0.01 | 0.07 | 0.02  |
| **W⁺** | Attack | -0.04 | -0.08 | 0.05 | -0.08 | 0.01 | -0.08 |

F: Formspring, T: Twitter, W: Wikipedia, ⁺: Oversampled

As results show, the performance of the models was influenced by different initial word embeddings. Using Random initial word embeddings outperformed the SSWE and GloVe in original datasets. However, in oversampled datasets, the initial word embeddings did not show a significant effect on cyberbullying detection. We noticed that the inconsistencies among the reference and reproduced *F1*-scores mostly occurred in CNN model on Twitter dataset (Table 2b).

The performance of all four DNN models while using SSWE as the initial word embeddings is summarized in Table 3a. Same as the reference work, we also noticed that the LSTM were outperformed by other models, and the performance gaps in the other three models were quite small. Following the above mentioned inconsistencies, we further observed that the main difference between the reproduced and reference results are due to the differences in recall values (Table 3b).

**Table 3a.** Performance comparison of the various DNN models with SSWE as the initial word embeddings - reproduced

| Dataset | Label  | Precision |      |       |                   | Recall |      |       |                   | *F1*-score |      |       |                   |
|---------|--------|-----------|------|-------|-------------------|--------|------|-------|-------------------|------------|------|-------|-------------------|
|         |        | CNN       | LSTM | BLSTM | BLSTM attention   | CNN    | LSTM | BLSTM | BLSTM attention   | CNN        | LSTM | BLSTM | BLSTM attention   |
| **F+**  | Bully  | 0.94      | 0.78 | 0.79  | 0.86              | 0.61   | 0.93 | 0.94  | 0.92              | 0.74       | 0.85 | 0.76  | 0.89              |
| **T+**  | Racism | 0.97      | 0.96 | 0.97  | 0.96              | 0.77   | 0.92 | 0.99  | 0.99              | 0.86       | 0.94 | 0.98  | 0.97              |
|         | Sexism | 0.94      | 0.89 | 0.95  | 0.94              | 0.60   | 0.96 | 0.99  | 0.99              | 0.73       | 0.92 | 0.97  | 0.97              |
| **W+**  | Attack | 0.93      | 0.90 | 0.94  | 0.93              | 0.82   | 0.91 | 0.98  | 0.98              | 0.87       | 0.91 | 0.96  | 0.96              |

F: Formspring, T: Twitter, W: Wikipedia, ⁺: Oversampled

**Table 3b.** Difference in the performance of the reference and the reproduced DNN models with SSWE as the initial word embeddings. Negative values indicate higher performance of the reproduced models.

| Dataset | Label | Precision |      |       |                 | Recall |      |       |                 | *F1*-score |      |       |                 |
|---------|-------|-----------|------|-------|-----------------|--------|------|-------|-----------------|------------|------|-------|-----------------|
|         |       | CNN       | LSTM | BLSTM | BLSTM attention | CNN    | LSTM | BLSTM | BLSTM attention | CNN        | LSTM | BLSTM | BLSTM attention |

| | | | | | | | | | | | | |
|---|---|---|---|---|---|---|---|---|---|---|---|---|
| **F+** | Bully | -0.01 | 0.13 | 0.12 | 0.04 | 0.29 | -0.08 | -0.13 | -0.01 | 0.17 | 0.03 | 0.1 | 0.02 |
| **T+** | Racism | -0.04 | -0.05 | -0.05 | -0.06 | 0.17 | -0.12 | -0.04 | -0.03 | 0.07 | -0.09 | -0.05 | -0.04 |
| | Sexism | -0.02 | -0.05 | -0.07 | -0.06 | 0.32 | -0.03 | -0.05 | -0.07 | 0.19 | -0.04 | -0.05 | -0.07 |
| **W+** | Attack | -0.01 | -0.20 | -0.04 | -0.06 | 0.01 | -0.37 | -0.17 | -0.12 | 0.01 | -0.30 | -0.11 | -0.09 |

F: Formspring, T: Twitter, W: Wikipedia, +: Oversampled

## 3    Application and Evaluation of the Methods on a New Dataset

In this section we investigated the adaptability of the reproduced methods on a new dataset and to evaluate the performance of the methods on a new social media platform; YouTube. We investigate how the DNN models would perform in this dataset in comparison to the previous ML models used on this dataset for detection of cyberbullying.

### 3.1    YouTube Dataset

YouTube is one of the most popular user-generated content video platforms. The wide range of audience and the content of the videos make it prone to misbehaviours such as cyberbullying. In this study we use a YouTube dataset which has been created by [6]. This dataset was developed by searching the YouTube for topics sensitive to cyberbullying, such as, gender and race. From the retrieved videos, the comments of the users as well as their publicly available profile information were extracted. After some trimming, the final dataset consists of about 54,000 comments from 3,858 distinct users. The comments were manually annotated by 2 persons. In total about 6,500 of the comments were labelled as bullying.

### 3.2    Workflow and Results

We used the same experimental settings as explained in Section 2 for the new dataset. We continued our experiments by implementing the DDN based models. The YouTube dataset also suffers from the class imbalance and the number of bullying posts is significantly smaller than the neutral ones. Therefore were oversampled the bullying posts of the dataset and their number was tripled. Table 4 shows the performance of the BLSTM with attention model using the three initial word embeddings in both the original dataset and the oversampled dataset. We also used different dimension sizes for word embeddings, from 25 to 200. Here we have reported the average of all the experimented demission sizes for each word embedding.

**Table 4.** Performance evaluation of the BLSTM with attention models, with three word embedding methods on the YouTube original dataset and the oversamples dataset.

| **Dataset** | **Label** | **Precision** | **Recall** | *F1*-score |
|---|---|---|---|---|

Dadvar and Eckert 9

|   |       | Random | Glove | SSWE | Random | Glove | SSWE | Random | Glove | SSWE |
|---|-------|--------|-------|------|--------|-------|------|--------|-------|------|
| **Y** | Bully | 0.41 | 0.47 | 0.19 | 0.18 | 0.29 | 0.11 | 0.22 | 0.35 | 0.12 |
| **Y⁺** | Bully | 0.93 | 0.89 | 0.96 | 0.91 | 0.84 | 0.88 | 0.92 | 0.86 | 0.92 |

Y: YouTube, ⁺: Oversampled

As the results show, oversampling of the dataset significantly (Mann-Whitney U test, P<0.001) improved the performance of the models in all three word embeddings. Overall, the SSWE has the highest *F1*-score and precision while Random embeddings resulted in the highest recall. The performance measures of all the DNN based models using SSWE initial word embedding is presented in Table 5. The LSTM model had the lowest performance in comparison to other models, mainly due to near zero recall. While BLSTM has the highest *F1*-score and recall, BLSTM with attention also performed quite similar with slightly higher precision in the cost of a lower recall.

**Table 5.** Performance comparison of the various DNN models with SSWE as the initial word embeddings – YouTube Data

| Dataset | Label | Precision | | | | Recall | | | | *F1*-score | | | |
|---------|-------|-----------|---|---|---|--------|---|---|---|------------|---|---|---|
|         |       | CNN | LSTM | BLSTM | BLSTM attention | CNN | LSTM | BLSTM | BLSTM attention | CNN | LSTM | BLSTM | BLSTM attention |
| **Y⁺** | Bully | 0.88 | 0.86 | 0.94 | 0.96 | 0.7 | 0.08 | 0.93 | 0.88 | 0.78 | 0.14 | 0.93 | 0.92 |

Y: YouTube, ⁺: Oversampled

### 3.3 Transfer Learning

Transfer learning is the process of using a model which has been trained on one task for another related task. Following [11] we also implemented the transfer learning procedure to evaluate to what extent the DNN models trained on a social network, here Twitter, Formspring, and Wiki, can successfully detect cyberbullying posts in another social network, i.e., YouTube. For this purpose we used the BLSTM with attention model and experimented with three different approaches.

- *Complete Transfer Learning.* In this approach, a model trained on one dataset is directly used in other datasets without any extra training. As the results in Table 6 show, the recalls are quite low but varying in all three datasets. This can indicate that the nature of cyberbullying is different in these different datasets. However, the complete transfer learning approach shows that bullying nature in YouTube is more similar to Formspring (*F1*-score = 0.30) and then to Wikipedia (*F1*-score = 0.23) in comparison to Twitter (*F1*-score = 0.15). This might be due to the similarity of the nature of these social networks. YouTube, Formspring and Wikipedia all have longer posts and are task oriented, while Twitter has short posts and is more general purpose.
- *Feature Level Transfer Learning.* In this approach, a model is trained on one dataset and only learned word embeddings are transferred to another dataset for training a new model. The evaluation metrics of the transferred models show improvements (Table 6) in both precision and recall for all the datasets. This improvement compared to previous leading approach, indicates the importance of learned word embeddings and their impact cyberbullying detection. As illustrated in the table, Wikipedia has the highest performance with $F1$-score = 0.74.
- *Model Level Transfer Learning.* In this approach, a model is trained on one dataset and learned word embeddings, as well as network weights, are transferred to another dataset for training a new model. The improvement of results in this learning approach (Table 6) was not significant compared to the feature level learning approach. This indicates that the transfer of network weights is not as essential to cyberbullying detection as the learned word embeddings.

**Table 6.** Performance comparison of the three transfer learning approaches using BLSTM with attention

| Train \ Test Dataset | Approaches | Y Precision | Y Recall | Y $F1$-score |
|---|---|---|---|---|
| **F+** | Complete | 0.74 | 0.19 | 0.30 |
| | Feature Level | 0.82 | 0.50 | 0.62 |
| | Model Level | 0.73 | 0.80 | 0.76 |
| **T+** | Complete | 0.46 | 0.09 | 0.15 |
| | Feature Level | 0.89 | 0.52 | 0.66 |
| | Model Level | 0.83 | 0.99 | 0.90 |
| **W+** | Complete | 0.68 | 0.14 | 0.23 |
| | Feature Level | 0.81 | 0.67 | 0.74 |
| | Model Level | 0.99 | 0.96 | 0.97 |

F: Formspring, T: Twitter, W: Wikipedia, Y: YouTube, +: Oversampled

## 4    Conclusion and Future Work

In this study, we successfully reproduced the reference literature [11] for detection of cyberbullying incidents in social media platforms using DNN based models. The source codes and materials were mostly well organized and accessible. However, there were some details and settings that were not clearly stated. These might have been the reason for some the inconsistencies in our results. We further expanded our work by using a new social media dataset, YouTube, to investigate the adaptability and transferability of the models to the new dataset and also to compare the performance of the DNN models against the conventional ML models which were used in previous studies on the YouTube dataset for cyberbullying detection.

Often, the datasets for cyberbullying detections contains very few posts marked as bullying. This imbalance problem can be partly covered by oversampling the bullying posts. However, the effects of such prevalence on the performance of models need to be further assessed. Our study shows that the DNN models were adaptable and transferable to the new dataset. DNN based models coupled with transfer learning outperformed all the previous results for the detection of cyberbullying in this YouTube dataset using ML models. In [26], [27] authors have used context-based features such as the users' profile information and personal demographics to train the ML models which has resulted in F1-score=0.64. In [6] the discrimination capacity of the detection methods were improved to 0.76 by incorporating expert knowledge.

We believe that the DNN models can also benefit from integrating other sources of information and as the future work we recommend to look into the impact of profile information of the social media users and to investigate the improvement of the models by considering the above mentioned sources of information.

**Acknowledgment.** The authors gratefully acknowledge the kind support of Sweta Agrawal in the reproduction of the work. They are also thankful to Aidin Niamir and the Senckenberg Data and Modelling Centre for their kind support with high performance computing. MD is a postdoctoral researcher in Fachinformationsdienst Judaica project, funded by Deutsche Forschungsgemeinschaft (DFG 286004564).